\documentclass[runningheads]{llncs}
\usepackage[T1]{fontenc}
\usepackage{graphicx,verbatim}
\usepackage{tcolorbox}
\usepackage{xcolor}
\usepackage{hyperref}
\newcommand{\DIWEI}[1]{\textcolor{red}{#1}}

\begin{document}
\title{AGIR: Assessing 3D Gait Impairment with Reasoning based on LLMs 
}

%

\author{Anonymized Authors}  

\author{Diwei Wang\inst{1}
\and
Cédric Bobenrieth\inst{1,2} 
\and Hyewon Seo \inst{1}
}
\authorrunning{D. Wang et al.}
\institute{ICube laboratory, University of Strasbourg, CNRS, France
\\
\email{\{d.wang, seo\}@unistra.fr}\\
\and ICAM Strasbourg Europe, France \\
\email{\{cedric.bobenrieth\}@icam.fr}\\
}    
\maketitle              
\begin{abstract}
Assessing gait impairment plays an important role in early diagnosis, disease monitoring, and treatment evaluation for neurodegenerative diseases. Despite its widespread use in clinical practice, it is limited by subjectivity and a lack of precision. While recent deep learning-based approaches have consistently improved classification accuracies, they often lack interpretability, hindering their utility in clinical decision-making. To overcome these challenges, we introduce AGIR, a novel pipeline consisting of a pre-trained VQ-VAE motion tokenizer and a subsequent Large Language Model (LLM) fine-tuned over pairs of motion tokens and Chain-of-Thought (CoT) reasonings. To fine-tune an LLM for pathological gait analysis, we first introduce a multimodal dataset by adding rationales dedicated to MDS-UPDRS gait score assessment to an existing PD gait dataset. We then introduce a two-stage supervised fine-tuning (SFT) strategy to enhance the LLM's motion comprehension with pathology-specific knowledge. This strategy includes: 1) a generative stage that aligns gait motions with analytic descriptions through bidirectional motion-description generation;  2) a reasoning stage that integrates logical Chain-of-Thought (CoT) reasoning for impairment assessment with UPDRS gait score. Validation on an existing dataset and comparisons with state-of-the-art methods confirm the robustness and accuracy of our pipeline, demonstrating its ability to assign gait impairment scores from motion input with clinically meaningful rationales. Our dataset, code, and demo videos are accessible on our project page at: \url{https://anonymous.4open.science/w/AGIR-7BF7/}.
\keywords{MDS-UPDRS Gait Score Estimation\and Large Language Models \and Chain-of-thought Reasoning}

\end{abstract}

\section{Introduction}
Gait analysis serves as a non-invasive, objective, and cost-effective method for early diagnosis, disease monitoring, and treatment evaluation in neurodegenerative diseases, 
such as Parkinson’s disease (PD), Alzheimer’s disease (AD), and Dementia with Lewy Bodies (DLB). 
MDS-UPDRS (Movement Disorder Society's Unified Parkinson’s Disease Rating Scale), the most widely  used clinical scaling system for assessing the severity and progression of Parkinson's disease (PD), classifies the patient's walking ability into five levels, ranging from 0: Normal to 4: Severe difficulty.

There is growing interest in developing automatic scoring methods for MDS-UPDRS to reduce subjectivity, enhance consistency, and improve assessment efficiency. These methods leverage advanced technologies such as wearable sensors, machine learning algorithms, and computer vision to objectively measure and score gait motions. 
In particular, recent developments in deep learning techniques have shown notable performance in computational gait classification. Methods deploying Temporal Convolutional Neural Network (TCNN) \cite{lu2020miccai}, Spatio-Temporal Graph Convolutional Network (ST-GCN)  \cite{sabo2022estimating}, or Transformer \cite{max-gr2023} have achieved state-of-the-art performances. Other approaches deploying a Vision Language Model (VLM) \cite{Wang_MICCAI2024} have also shown promising results, with the possibility of decoding the vision-text aligned embedding into textual descriptions that include gait parameters. 
While these analyses are often framed as classification problems, the justification of classification results, along with a descriptive analysis of the given gait motion, remains largely unexplored. Our work aims to address this issue by developing an advanced model that provides diagnostic rational or reasoning about gait motion in addition to the class predictions.

The powerful abilities of large language models have enabled detailed and contextually relevant interpretations of medical images, supporting applications such as automated medical report generation \cite{insight_llm2024}, disease diagnosis from multimodal data \cite{LLM_LungCancer}, and visual question answering \cite{PitVQA}.
MedBLIP \cite{wang2022medclip} adapts BLIP (Bootstrapping Language-Image Pre-training) framework to the medical domain, improving the accuracy and efficiency of medical image analysis by integrating visual and textual information.
\cite{LLM_LungCancer} employs an LLM to align CT images, pathology slides, and clinical notes for predicting the 5-year survival of lung cancer patients. Insight \cite{insight_llm2024} introduces an LLM-based clinical report summarizer, which contextualizes clinical metadata and image-driven morphology data in a Q\&A form. 
Inspired by these successes, we propose to leverage large language models as an informative evaluator for gait impairment analysis, 
offering interpretable reasoning behind each assessment score to improve understanding and reliability in clinical evaluations.
\begin{figure}[!ht]
\includegraphics[width=0.98\textwidth]{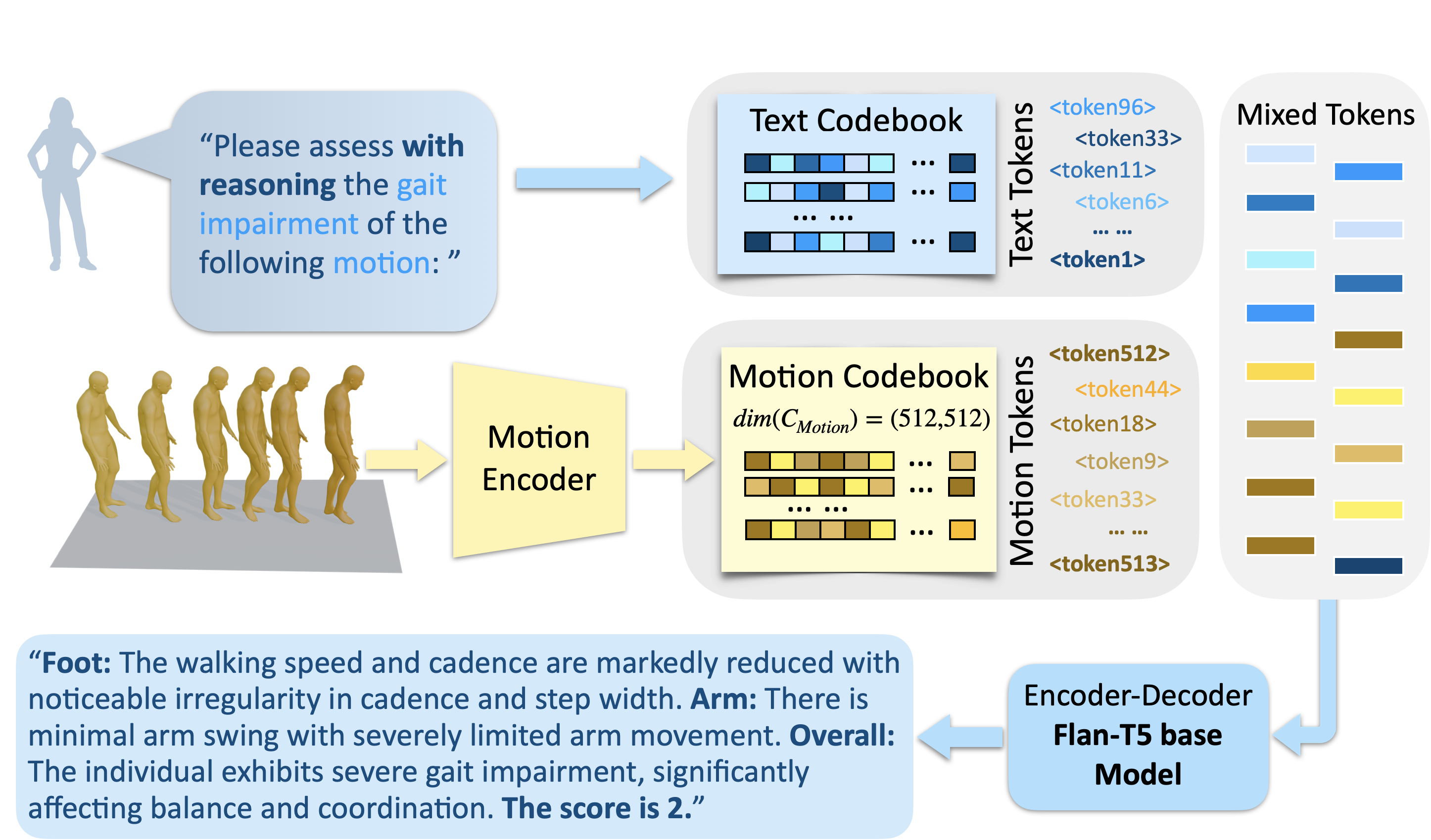}
\caption{Our approach, AGIR, estimates an MDS-UPDRS \cite{mds-updrs2008} gait impairment score and provides an analytical description of the input gait motion. 
} 
\label{teaser}
\end{figure}
Interestingly, 
some studies show that human motion can be represented as a form of language by quantizing motion signals into discrete tokens, with VQ-VAE \cite{vq-vae} being a widely used framework for this purpose. TM2T \cite{Guo2022tm2t} discretized and compressed raw motions into tokens, aligning them with text tokens. 
The motion2text module was jointly trained with text2motion, employing a cross-modal reconstruction loss to penalize significant deviations of synthesized text from the input text.
MotionGPT \cite{jiang2023motiongpt} introduces a pre-trained motion-language model capable of generalizing to various motion-related downstream tasks using prompting, aiming to model the relation between language and motion within a unified multi-task framework. However, improving the accuracy and efficiency of these models within the clinical domain remains yet to be addressed.
A recent LLM-based action recognizer \cite{qu2024llms} demonstrates the effectiveness of LLMs in 3D action recognition tasks. The representation learning of the VQ-VAE model \cite{vq-vae} optimized for action signals is followed by the fine-tuning of a pre-trained LLM using low-rank adaptation (LoRA) with action tokens derived from raw skeleton data.
However, it does not provide reasonings for the classifications made by the model.

In this paper, we design an LLM-based assessment for gait impairment that is able to provide diagnostic rationale beyond simple ratings. As Fig. \ref{teaser} illustrates, our approach, named AGIR, delivers both precise classification and interpretable reasoning for gait impairments in the given motion by encoding and aligning the motion data with descriptive textual information.
To the best of our knowledge, we are the first to integrate an LLM with 
gait data for interpretable pathological gait analysis. 
Specifically, we make the following contributions:
\begin{itemize}
\item We enhance a publicly available 3D pathological gait motion dataset by annotating the motion data with analytical text associated to the MDS-UPDRS gait score. This makes the dataset readily available for multimodal learning, allowing us to achieve our goal.
\item We establish a new paradigm for the assessment of gait impairment by defining a novel objective that combines reasoning alignment with score classification, enabling interpretable assessments through gait specific analytical descriptions.
\item We validate the model performance on our multimodal dataset using both classification and semantic alignment metrics, showcasing the efficacy of the proposed strategy.
\end{itemize}
\section{Method}
\subsection{Multimodal gait dataset for impairment assessment}
\begin{figure}
\includegraphics[width=0.98\textwidth]{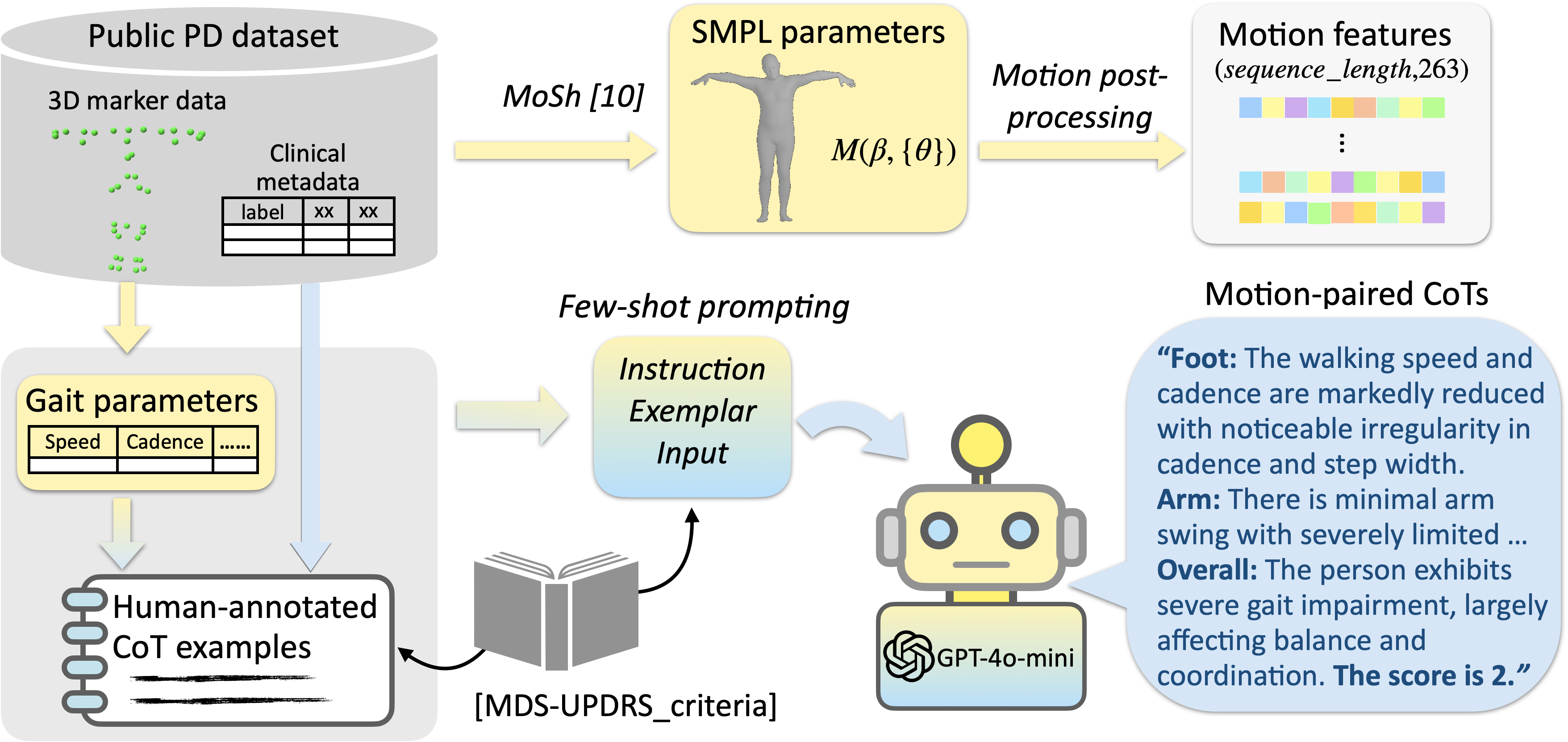}
\caption{Pipeline for our multimodal dataset preparation.}
\label{dataset}
\end{figure}

We enhance an existing Parkinson's Disease (PD) gait dataset \cite{public2023dataset} by incorporating analytical descriptions related to clinical motor symptom evaluations. This public dataset contains kinematic data from static and gait trials of 26 idiopathic Parkinson's disease patients in ON and OFF medication conditions, captured with 44 anatomical reflective markers. It also includes metadata on patient demographics and clinical assessments based on MDS-UPDRS \cite{mds-updrs2008}. To adapt the dataset for LLM-based multimodal learning, we introduce a pipeline that 
produces motion data paired with corresponding gait scoring rationales, as detailed below. Fig.~\ref{dataset} illustrates this data preparation pipeline.
\subsubsection{Motion data processing.}
We reformat the motion data into the HumanML3D format \cite{guo2022humanml3d}, making it compatible with the pre-trained Vector Quantised-Variational Autoencoder (VQ-VAE) motion tokenizer. 
It represents motion as time-series data of \textit{motion features}---3D joint positions, velocities, rotations, and foot-contact indicators. 
We begin by downsampling the high-resolution motion data from 100 fps to 20 fps. 
Next, we transform the marker-based motion data into joint angle-based SMPL representations \cite{loper2015smpl} by applying the MoSh method \cite{loper2014mosh}. The SMPL model represents a 3D human body in motion using a shape parameter $\beta$ and a sequence of pose parameters \{$\theta$\}: $M(\beta, \{\theta\})$. We used the static trial of each subject to estimate a shape parameter $\beta$, which was then used consistently when computing the pose parameters $\{\theta\}$'s across all gait trials of the same subject. Trials in which Mosh fails to convert have been excluded, resulting in a final dataset of 883 walking sequences from 26 subjects. 
Finally, we post-processed the SMPL representation to extract motion features that are compatible with the HumanML3D format.
\subsubsection{Chain-of-Thought (CoT) annotation preparation.}\label{CoT prepare}
Although previous works, such as MotionGPT \cite{jiang2023motiongpt}, enable the automatic synthesis of text descriptions for given 3D human motions, they lack diagnostic reasoning capabilities, making them unsuitable for our purpose.
Thus, we develop a pipeline to annotate each gait motion with a sequence-level analytical description, 
which represent the interpretation of the provided gait score. This is achieved by few-shot prompting the GPT-4o-mini model \cite{hurst2024gpt}
with an \textit{Instruction-Exemplar-Input} format:
\begin{tcolorbox}
\scriptsize { 
  \textit{\textbf{Instruction.}} Given a set of gait parameters, you will give qualitative reasoning for the provided gait score with reference to the gait score assessment criteria [MDS-UPDRS\_citeria], and in reference to gait parameters of the normal elderly [normal\_reference].\\
\textit{\textbf{Exemplars.}}\\\textit{**Question**  }The MDS-UPDRS Gait score is 0. Cadence: 110.0, Speed: 1.26m/s, Coefficient of variation of speed: 3.4\%, ......, Maximal vertical height of the foot from the ground in swing phase: 0.17m.\\
\textit{**Answer**\quad}Foot: The walking exhibits a consistent and regular step length and height, contributing to a well-balanced gait. Arm: Arm movements are well-coordinated with foot movements, facilitating stability and rhythmic motion. Overall: The overall gait pattern is stable, rhythmic, and efficient.\\
\textit{**Question** }The MDS-UPDRS Gait score is 1. ......\\
\textit{**Answer**\quad}Foot: ...... Arm: ...... Overall: ......
\\
\textit{\textbf{Input question.}} The MDS-UPDRS Gait score is 2. Cadence: 46.0, Speed: 0.18 m/s, Coefficient of variation of speed: 37.7\%, ......, Maximal vertical height of the foot from the ground in swing phase: 0.07m.
}
\end{tcolorbox}
\noindent As shown above, the prompt incorporates the MDS-UPDRS gait score from the clinical metadata, CoT examples (9 total, 3 per class), as well as the gait parameters extracted from the motion. The CoT examples formulate scoring rationales for the motion, referencing the mean gait parameters of normal walking and per-class criteria. Drawing inspiration from \cite{adeli2024benchmarking}, we select a set of gait parameters that are highly  relevant for UPDRS gait score estimation, 
such as cadence, speed, average step length, arm swing angles, and margin of stability. 
The complete list of parameters along with their computation details are provided on our project webpage.
We revised the generated CoTs manually to correct potential errors.

\subsection{LLM supervised fine-tuning for explainable gait analysis} 
We perform a two-stage supervised fine-tuning (SFT) on the Flan-T5 base model, which is pretrained by MotionGPT \cite{jiang2023motiongpt}:
\begin{enumerate}
    \item \textit{SFT for bidirectional motion-text generation (Generative stage)}.
We inform the pretrained language model with gait-specific knowledge by minimizing  the token-level reconstruction error of a motion-description pair, using a randomly selected modality as input. 
We use the foot and arm descriptions as the motion-paired text for learning, as shown in examples in Sec~\ref{CoT prepare}. 
    
\item \textit{SFT with instructions for explainable gait score prediction (Reasoning stage)}. We prompt the model with instructions to elicit the reasoning of the language model. To enhance training diversity, we utilized 200 different instruction templates.
\end{enumerate}

\begin{figure}
\includegraphics[width=0.98\textwidth]{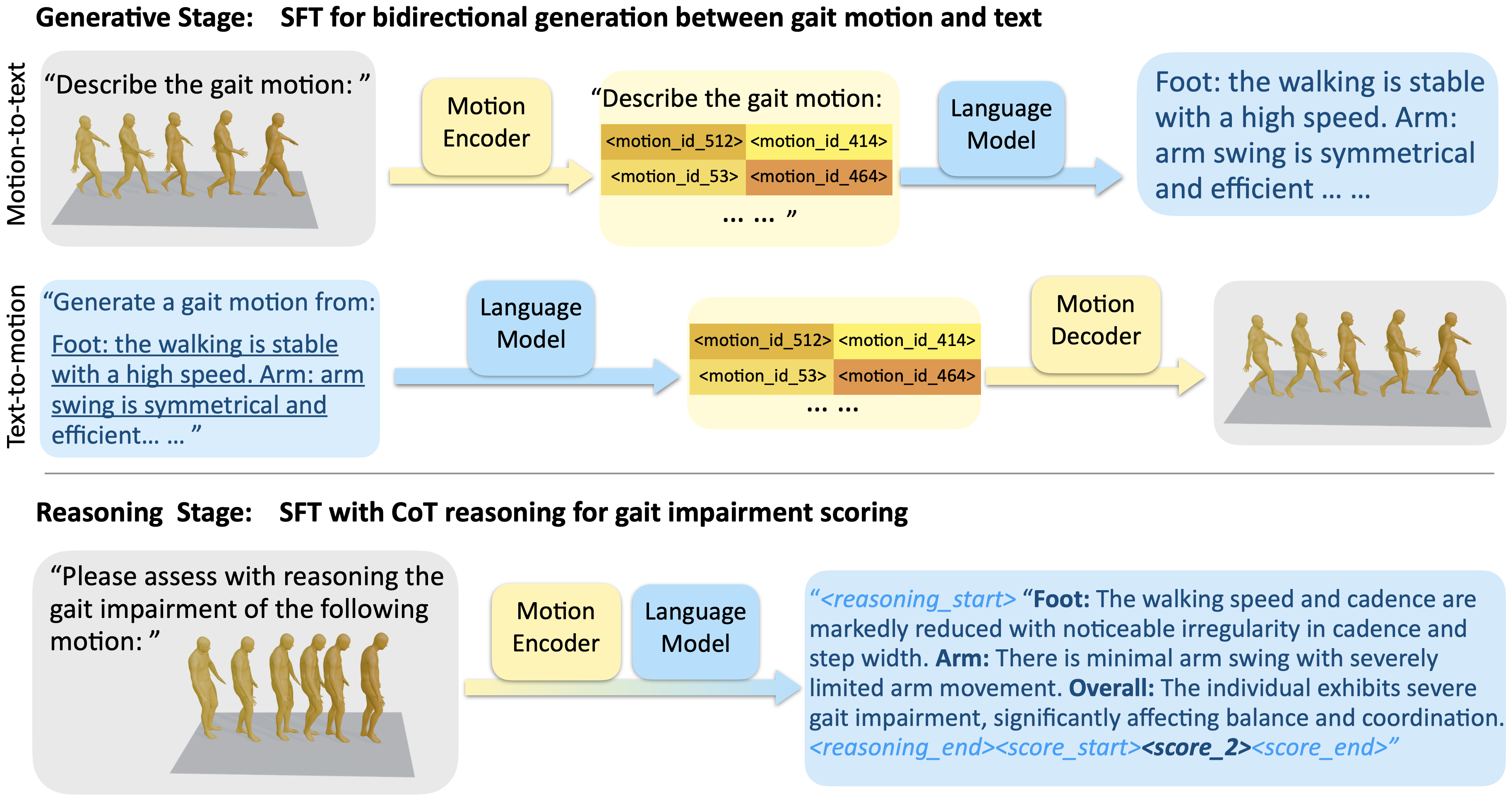}
\caption{Overview of our 2-stage supervised fine-tuning framework.} \label{scheme}
\end{figure}

As illustrated in Fig.~\ref{scheme}, our model is designed to generate text outputs adhering to a predefined structure for the reasoning task. To achieve this, seven special tokens were integrated into the pre-trained Flan-T5 model, increasing its vocabulary size by 7. Specifically, <$reasoning\_start$> and <$reasoning\_end$> denote the boundaries of the gait score explanation, while <$score\_start$> and <$score\_end$> indicate the assessment result. Additionally, <$score\_0$>, <$score\_1$>, and <$score\_2$> represent progressively increasing levels of symptom severity, 
as defined in the MDS-UPDRS III Gait criteria. To enforce the structured text output of the model for the reasoning task, we employed a format loss based on the Sigmoid function $\sigma$:     $L_{fmt}=\sum^{5}_{k=1}(1-\sigma(\overline{Z} [pos_{k}, T_k]))$,
where $\overline{Z}$ is the sequence of output vectors from our model, normalized using a LogSumExp (LSE) function: $\overline{Z}=Z-LSE(Z)$. $k$ corresponds to the positions of 5 special tokens required in the structured output, while $T_k$ denotes their respective token IDs. 
To determine the positions of special tokens in a structured text output, we first enforce the prediction of <$reasoning\_start$> at the beginning of the text and identify the <$eos$> token, which marks the end of the sentence. The remaining 4 special tokens are then constrained to appear sequentially before <$eos$>. The loss function $L_{fmt}$ maximizes the likelihood of the positional indicator tokens, thereby simultaneously optimizing the prediction of the <$score\_[label]$> for gait score classification.

Additionally, we used a standard language modeling loss $L_{LM}$ which maximizes the likelihood of the correct text generation with a cross-entropy objective.  We also integrated another cross-entropy loss, $L_{cls}$, on the score token prediction to explicitly penalize classification errors. 

\section{Experiments and Results}
\subsubsection{Comparison of general and our pathological motion encoders.}
We validate our approach by comparing the motion embeddings fine-tuned through the 2-stage SFT with those generated by MotionGPT, which uses the same VQ-VAE motion tokenizer.
Following a similar experimental setting of a recently proposed benchmark \cite{adeli2024benchmarking}, we employed a single-layer linear classifier with layer normalization to predict gait score from the average motion token embeddings of each gait motion. We further trimmed the motion into clips of 54 frames with a stride length of 27 frames to train the MLP. During testing, we computed the average motion token embeddings to predict classification results on fixed-length sequences, followed by majority voting to generate per-motion gait scores.

The results shown in Table \ref{sota} compares the classification results of state-of-the-art motion encoders from the benchmark \cite{adeli2024benchmarking} on Leave-One-Subject-Out-Cross-Validation. Notably, 
our fine-tuning strategy effectively optimized motion token embeddings, improving task-specific gait classification performance over state-of-the-art motion encoders.
We also observed that integrating a generative stage in SFT contributes to significant improvement compared with a single reasoning stage, validating the effectiveness of 2-stage SFT. 

\begin{table}[!ht]
\centering
\caption{Evaluation of the motion embedding learned through 2-stage SFT training on gait classification against the state-of-the-art 
methods reported in the benchmark \cite{adeli2024benchmarking}. 
Classification performance is measured with top-1 accuracy(\%), recall rate(\%), precision rate(\%), and F1-Score. We highlight the best performances in \textbf{boldface}.}\label{sota}
\begin{tabular}{|l |c|c|c|c|}
\hline
Configuration &\multicolumn{4}{c|}{Gait score classification} \\
\cline{2-5}& Accuracy & Recall\quad & Precision & F1-Score\\
\hline
MotionAGFormer \cite{mehraban2024motionagformer} &0.42 &0.42 &0.42 &0.42 \\ 
PD STGCN \cite{sabo2022estimating} &0.49 &0.50 &0.49 &0.48\\
PoseFormerV2 \cite{zhao2023poseformerv2} &0.49 &0.50 &\textbf{0.49} &0.48\\
\hline
\hline
VQ-VAE+MotionGPT~\cite{jiang2023motiongpt} &0.44 &0.44 &0.39 &0.58\\
VQ-VAE+Reasoning stage only &0.49 &0.49 &0.38 &0.30\\
VQ-VAE+2-stage SFT training (ours) &\textbf{0.51} &\textbf{0.51} &0.46 &\textbf{0.58} \\
\hline
\end{tabular}
\end{table}
\subsubsection{Ablation studies.}
In Table \ref{ablation}, we study the impact of various elements in our pipeline on several metrics to evaluate the quality of the generated text compared to reference outputs. The metrics chosen, commonly used in natural language processing tasks, include CIDEr \cite{krishna2015cider}, and Bert-F1 \cite{zhang2020bertscore}.
The elements considered are single-stage (reasoning only), mixed stage (i.e. reasoning with embedded generative objective (EGO)) and two-stage SFT, which is our design choice. Given that rationales for gait scoring may lack the explicit logical flow typically found in reasoning chains for tasks like math word problems, we have examined the integration of Direct Preference Optimization (DPO) \cite{rafailov2023dpo} to strengthen the causal relationship between reasoning and the resulting gait score. During training, we constructed a triplet $(X, R_w, R_l)$ for a sampled motion $X$, with $R_w$ as the preferred CoT and $R_l$ as the undesired one. Typically, we curated both irrelevant and counterfactual CoT as $R_l$, following \cite{making2024reasoning}. However, the inclusion of DPO in the form of a classification loss $L_{DPO}$ as proposed in \cite{making2024reasoning} did not show a significant impact. This observation somehow aligns with the findings in Smaug \cite{pal2024smaug}, where low edit distances between $R_w$ and $R_l$ leads to reduced prediction performance. Since our reasoning annotations share limited vocabularies constrained by the UPDRS assessment criteria, the edit distances between $R_w$ and $R_l$ remain relatively low.

We conducted a standard cross-validation on our multi-modal dataset. For each configuration in Table~\ref{ablation}, we evaluated the model with best CIDEr on the validation set. 
The results validate a synergized performance achieved with the inclusion of a bidirectional generative stage, enhancing both reasoning alignment and gait score classification. 
The motion-text generation objective also improves the reasoning alignment when embedded into the reasoning stage (EGO).

\begin{table}[!ht]
\caption{Evaluation of models with different configurations. Ground-truth reasoning alignment is assessed using CIDEr \cite{krishna2015cider} and Bert-F1 \cite{zhang2020bertscore}, while gait score classification performance is measured in top-1 accuracy(\%), precision rate(\%), recall rate(\%), and F1-Score. In the first column, EGO stands for embedded generative objective. We highlight the best performance for each metric in \textbf{boldface}.}\label{ablation}
\begin{tabular}{|l|c|c|c|c|c|c|}
\hline
Configuration &\multicolumn{2}{c|}{Reasoning}&\multicolumn{4}{c|}{Gait score classification} \\
\cline{2-7}&CIDEr $\uparrow$ &Bert-F1 $\uparrow$ &acc. &recall &prec. &F1-score\\
\hline
Reasoning stage only &0.78 &0.80 &0.45 &\textbf{0.47} &\textbf{0.66} &\textbf{0.49}\\
Reasoning stage with EGO &0.88 &0.82 &0.49 &0.45 &0.51 &0.44\\
Reasoning stage with $L_{DPO}$ &0.93 &0.82 &0.38 &0.34 &0.37 &0.33\\
Reasoning stage with EGO \& $L_{DPO}$ &0.92 &0.82 &0.37 &0.46 &0.38 &0.36\\
Generative+Reasoning stages (AGIR) &\textbf{0.95} &\textbf{0.83} &\textbf{0.49} &0.45 &0.65 &0.45\\
\hline
\end{tabular}
\end{table}
\section{Discussion and Conclusion}
In this paper, we introduced a multimodal approach to pathological gait analysis, leveraging an LLM to integrate multifaceted gait motion data. We designed a two-stage instruction tuning strategy to fine-tune the LLM on our multimodal gait dataset, equipping it with domain-specific knowledge for gait analysis. Our approach enhances the LLM's ability to provide reasoned assessments of gait impairments, demonstrating its potential for clinical applications.
Experimental results demonstrate that our strategy not only improves diagnostic accuracy but also delivers robust and interpretable reasoning.
We  primarily focused on the UPDRS gait assessment in this study. 
In the future, we plan to extend our work to broader clinical domains by integrating evaluations for medication-induced movement abnormalities, such as levodopa-induced chorea in Parkinson's Disease.
\bibliographystyle{splncs04}\bibliography{MICCAI2025}

\end{document}